\documentclass{article}
\usepackage{spconf,amsmath,epsfig}
\usepackage{graphicx}
\usepackage{amssymb}
\usepackage{comment}
\usepackage{siunitx}
\usepackage{xcolor}

\usepackage{todonotes}

\begin{document}\sloppy

\def\x{{\mathbf x}}
\def\L{{\cal L}}

\title{Vertex Feature Encoding and Hierarchical Temporal Modeling in a Spatial-Temporal Graph Convolutional Network for Action Recognition}
%
\name{Konstantinos Papadopoulos, Enjie Ghorbel, Djamila Aouada, Bj\"{o}rn Ottersten}
\address{Interdisciplinary Centre for Security, Reliability and Trust (SnT)\\University of Luxembourg, Luxembourg\\  \small{\{\tt konstantinos.papadopoulos, enjie.ghorbel, djamila.aouada, bjorn.ottersten\}@uni.lu}}

\maketitle

\begin{abstract}
This paper extends the Spatial-Temporal Graph Convolutional Network (ST-GCN) for skeleton-based action recognition by introducing two novel modules, namely, the Graph Vertex Feature Encoder (GVFE) and the Dilated Hierarchical Temporal Convolutional Network (DH-TCN). On the one hand, the GVFE module learns appropriate vertex features for action recognition by encoding raw skeleton data into a new feature space. On the other hand, the DH-TCN module is capable of capturing both short-term and long-term temporal dependencies using a hierarchical dilated convolutional network. Experiments have been conducted on the challenging NTU RGB-D-60 and NTU RGB-D 120 datasets. The obtained results show that our method competes with state-of-the-art approaches while using a smaller number of layers and parameters; thus reducing the required training time and memory.
\end{abstract}

\section{Introduction}
\label{sec:introduction}
Skeleton-based human action recognition has received a huge amount of attention in various applications, such as video surveillance, coaching, and rehabilitation~\cite{6239233, ghorbel2018extension, 6909493, ghorbel2019view, papadopoulos2019two}. Recently, deep learning-based approaches have achieved impressive performance on large-scale datasets, by learning the appropriate features automatically from the data~\cite{10.1007/978-3-319-46478-7_23, caetano2019skelemotion, ke2018learning, liu2017enhanced, demisse2018pose, baptista2019view}. These approaches rely either on Recurrent Neural Networks (RNN) or Convolutional Neural Networks (CNN). However, they usually represent the skeleton sequences as vectors or 2D grids, ignoring inter-joint dependencies. 

To express joint correlations both spatially and temporally, Yan et al. introduced the Spatial Temporal-Graph Convolutional Network (ST-GCN)~\cite{yan2018spatial}. Their work takes advantage of Graph Convolutional Networks (GCN)~\cite{bruna2013spectral} extending the classical CNNs to graph convolutions. This architecture represents skeleton sequences as a graph composed of both temporal and spatial edges, by respectively considering the inter and intra-frame connections of joints. The effectiveness of this approach has motivated several extensions~\cite{shi2019two, shi2019skeleton, li2019actional} which, consider the most informative connections between joints instead of the predefined natural skeleton structure or construct the spatial-temporal graphs using additional features such as bone lengths.

However, all these methods only use raw skeleton features (joint coordinates and/or bone lengths) for the construction of spatial-temporal graphs. While offering a high-level description of the human body structure, these features may be lacking discriminative power for action recognition. Indeed, hand-crafted approaches have shown the limitation of using only raw skeleton joints as features in action recognition~\cite{zanfir2013moving, ghorbel2018kinematic}. Furthermore, the temporal dependencies of the graph are modeled by a single temporal convolutional layer. As a result, critical long-term dependencies might be not consistently described. Moreover, these approaches make use of a considerable number of ST-GCN blocks (10, in most cases), which significantly increases the number of parameters and consequently the computational complexity and the required memory.

In this paper, we assume that by encoding the vertex features in an end-to-end manner and modeling temporal long-term and short-term dependencies, less number of layers (and consequently parameters) will be needed. For that reason, two modules are introduced. The first module, referred to as Graph Vertex Feature Encoder (GVFE), is a trainable layer that transforms the feature space from the Euclidean coordinate system of joints to an end-to-end learned vertex feature space, optimized jointly with the ST-GCN. The second module incorporates a hierarchical structure of dilated temporal convolutional layers for modeling short-term and long-term temporal dependencies and replaces the standard temporal convolutional layers in the ST-GCN block. It is termed Dilated Hierarchical Temporal Graph Convolutional Network (DH-TCN). With the use of these two modules, we show that fewer layers are needed to reach the same or even higher performance in action recognition while needing less memory and training time than previous ST-GCN based approaches such as~\cite{li2019actional}.

In summary, our contributions are the following:
\begin{itemize}
    \item the introduction of a Graph Vertex Feature Encoder (GVFE) module for encoding vertex features;
    \item the proposal of a Dilated Hierarchical Temporal Graph Convolutional Network (DH-TCN) module for modeling short and long-term dependencies;
    \item the design of a more compact and efficient graph-based framework for action recognition trained in an end-to-end manner;
    \item the presentation of experimental validation and analysis of our approach on two challenging datasets.
\end{itemize}

\begin{figure*}[ht!]
\centering 
    \includegraphics[width=1\textwidth]{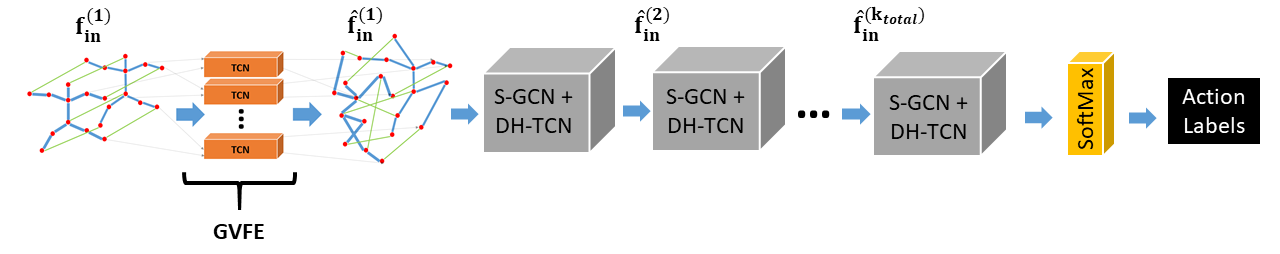}
    \caption{Illustration of the proposed approach. In the first step, the GVFE module generates graph features. The new graph is given as an input to the Modified ST-GCN blocks composed of a Spatial-Graph Convolutional Network (S-GCN) and a Dilated Hierarchical Temporal Convolutional Network (DH-TCN). Finally, a SoftMax layer classifies the spatial-temporal graph features resulting from the last Modified ST-GCN block.}
    \label{fig:pipeline}
\end{figure*}

The remainder of this paper is organized as follows: Section~\ref{sec:background} recalls the background related to Spatial-Temporal Graph Convolutional Networks (ST-GCN) applied to action recognition. Section~\ref{sec:pipeline} details the proposed framework. Section~\ref{sec:experiments} presents the experiments and analyzes the results. Finally, Section~\ref{sec:conclusion} concludes this paper and discusses possible extensions of this work.

\section{Background}
\label{sec:background}

\subsection{Skeleton Sequences as Graphs}


Following a predefined structure indicating their inter-connections, skeletons can be intuitively seen as graphs. Thus, Yan et al.~\cite{yan2018spatial} described skeleton sequences of $J$ joints and $T$ frames as spatial-temporal graphs in which, at each time instance $t$, each joint $i$ is assumed to be a vertex. Then, two kinds of edges are constructed to connect vertices: spatial edges that are the natural spatial joint connections and temporal edges connecting the same joint across time.  This spatial-temporal graph is denoted as $S=(V,E)$, with $V$ the set of vertices and $E$ the set of edges.

\subsection{Spatial-Temporal Graph Convolutional Network}

Considering the spatial-temporal graph $S$, a network called ST-GCN generalizing CNN to graphs has been proposed in~\cite{yan2018spatial}. 
In this work, for an input feature map $\mathbf{f_{in}}$, a spatial graph convolution is applied, such that: 

\begin{equation}
    \mathbf{f}_{out} = \mathbf{\Lambda}^{-\frac{1}{2}}(\mathbf{A}+\mathbf{I})\mathbf{\Lambda}^{-\frac{1}{2}} \mathbf{f}_{in} \mathbf{W},
\end{equation}

\noindent where $\mathbf{f_{out}}$ is the output feature map, $\mathbf{A}$ the adjacency matrix, $\mathbf{I}$ the identity matrix, $\mathbf \Lambda=[\Lambda^{ii}]_{i \in \{1,...,J\}}$ such that $\Lambda^{ii}=\sum_j(A^{ij}+I^{ij})$ and $\mathbf{W}$ is the weight matrix. For a graph of size $(C_{in},J,T)$, the dimension of the resulting tensor is $(C_{out},J,T)$, with $C_{in}$ and $C_{out}$ denoting respectively the number of input and output channels.

The temporal graph convolution consists of classical convolutions, performed on the output feature tensor $\mathbf{f}_{out}$. $K$ denotes the kernel size of the temporal convolutional layer.

The input features $\mathbf f_{in}^{(1)}$ incorporated in the first ST-GCN layer correspond to the joint coordinates such that $\forall i$, $\mathbf f_{in}^{(1)}(v_i,t)=\mathbf{P}_i(t)$ with $\mathbf{P}_i(t)$ the 3D coordinate of the joint $i$ at an instant $t$ and consequently $C_{in}=3$. While these first layer features $\mathbf f_{in}^{(1)}$  offer a representation easily understandable by the human, they might be not discriminative enough for the task of action recognition. Moreover, temporal graph convolutional layers capture only local dependencies, resulting in the presence of redundant information and neglecting long-term dependencies.

\section{Proposed Approach}
\label{sec:pipeline}

In this section, the two novel modules, namely GVFE and DH-TCN, are presented. While GVFE aims at learning vertex features, DH-TCN temporally summarizes spatial-temporal graphs and consequently models long-term as well as short-term dependencies. These two modules are integrated with the original ST-GCN~\cite{yan2018spatial} framework. This full pipeline is depicted in Fig.~\ref{fig:pipeline} and is trained in an end-to-end manner. It is important to note that these modules are also complementary to other ST-GCN extensions such as AS-GCN~\cite{li2019actional}.

\subsection{Graph Vertex Feature Encoding (GVFE)}
\label{sec:gvfe}

As mentioned in Section~\ref{sec:introduction}, considering raw skeleton joint data as vertex features might not be informative enough for action recognition. To enhance the discriminative power of vertex features, we introduce the GVFE module that is directly placed before the first ST-GCN block. GVFE maps 3D skeleton coordinates, traditionally used as input features to the first ST-GCN block $\mathbf f_{in}^{(1)}(v_i)=\mathbf P_i$ with $i \in \{1,...,J\}$, from the Cartesian coordinate system $\mathbb{R}^3$ to a learned feature space $\mathcal{M}\subseteq \mathbb{R}^{C_{out}}$. Since this module is trained in an end-to-end manner by optimizing the recognition error, we expect to obtain a more discriminative feature space $\mathcal{M}$.

\begin{figure}[t]
\centering 
    \includegraphics[width=0.5\textwidth]{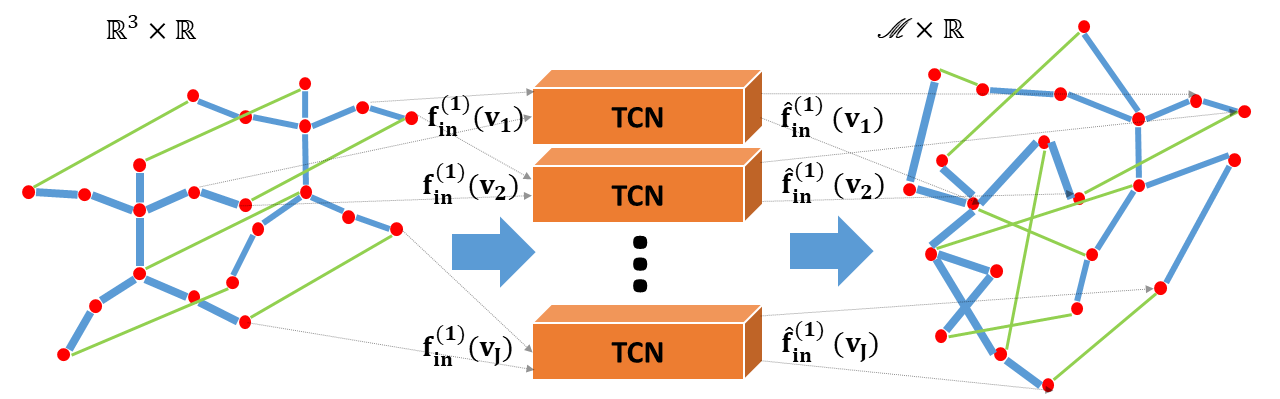}
    \caption{Illustration of the GVFE module structure: it is composed of $J$ TCN blocks. For each joint, one TCN block is separately used in order to conserve the natural skeleton structure.  }
    \label{fig:GVFE}
\end{figure}

For each joint $i$, a separate Temporal Convolutional Network (TCN) is employed to encode raw data, as illustrated in Fig.~\ref{fig:GVFE}. In this context, TCNs show strong potential since (a) they do not allow information to flow from the future states to the past states, (b) the input and output sequences have the same length and (c) they model temporal dependencies. For each joint $i$, the new graph vertex features $\mathbf{\hat {f}}_{in}^{(1)}(v_i)$ obtained after applying the TCN are computed as follows,
\begin{align}
    \mathbf{\hat {f}}_{in}^{(1)}(v_i) &= \mathbf{W}_i^{TCN} \ast  \mathbf f_{in}^{(1)}(v_i)= \mathbf{W}_i^{TCN} \ast \mathbf{P}_i,
\end{align}
\noindent where $\{\mathbf{W}^{TCN}_i\}_{1 \leq i \leq J}$ is the collection of tensors 
containing the kernel filters $\{\mathbf{W}^{TCN}_{i,j}\}$
of dimension $\mathbb R^{C_{out} \times T_w \times C_{in}}$, with $j \in \{1,...,C_{out}\}$ the index of the filter and $T_w$ the temporal size of the filters. Note that we use the identity activation function. 
This module has the advantage of being applicable to any graph-based network, regardless of the application.

\begin{figure}[t]
\centering 
    \includegraphics[width=0.45\textwidth]{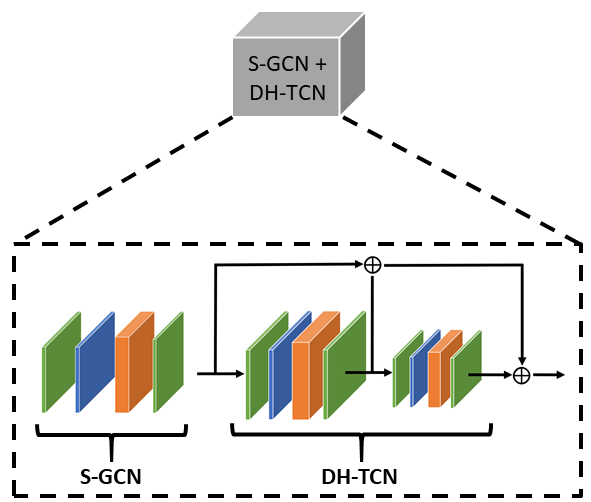}
    \caption{Illustration of S-GCN + DH-TCN block. Spatial features are extracted from the S-GCN module and are, then, fed into DH-TCN module. Green color is used for Batch Normalization units, blue for ReLU and orange for 2D Convolutional Layers.}
    \label{fig:block}
\end{figure}

\subsection{Dilated Hierarchical Temporal Graph Convolutional Network}
\label{sec:htcn}

The modeling of temporal dependencies is crucial in action recognition. However, in several ST-GCN-based approaches~\cite{yan2018spatial, shi2019two, shi2019skeleton, li2019actional}, temporal dependencies are modeled using only one convolutional layer. As a result, long-term dependencies that can be important for modeling actions are not well encoded. 

To that end, we propose to replace the temporal convolutions of each ST-GCN block with a module that encodes both short-term and long-term dependencies. Given the output feature map $\mathbf f^{(k)}_{out}$ resulting from the $k^{th}$ Spatial GCN (S-GCN) block (with $k \in [1, k_{total}]$ and $k_{total}$ the total number of ST-GCN blocks), this module, termed Dilated Hierarchical Temporal Convolutional Network (DH-TCN), is composed of $N$ successive dilated temporal convolutions. The association of these two blocks is illustrated in Fig.~\ref{fig:block}. Each layer output ${f}_{temp}^{(k,n)}$ of order $n$ of DH-TCN is obtained as follows,
\begin{equation}
    \mathbf{f}_{temp}^{(k,n)} = F \Big(\mathbf{W}_i^{DH} \ast_l  \mathbf f_{temp}^{(k,n-1)}\Big), \text{ with } \mathbf{f}_{temp}^{(k,0)} = \mathbf f^{(k)}_{out},
\end{equation}
\noindent where $\{ \mathbf W^{DH}\}_{1 \leq i \leq J}$ is the tensor containing the trainable temporal filters of dimension $\mathbb R^{C_{out} \times T_{w_1} \times C_{out}}$ with $T_{w_1}$ their temporal dimension and $\ast_l$ refers to the convolution operator with a dilation of $l=2^n, n\in [0,N-1]$.

The hierarchical architecture with different dilation ensures the modeling of long-term dependencies. At the same time, the residual connection depicted in Fig.~\ref{fig:block} enables the preservation of the information of short-term dependencies. 

The entire DH-TCN module is illustrated in Figure~\ref{fig:block}. Each hierarchical layer is composed of a dilated temporal convolution, a ReLU activation function, and a batch normalization.

\section{Experiments}
\label{sec:experiments}

Our framework has been tested on two well-known benchmarks, namely NTU RGB+D 60 (NTU-60)~\cite{Shahroudy_2016_NTURGBD} and NTU RGB+D 120 (NTU-120) \cite{Liu_2019_NTURGBD120} datasets. 

\subsection{Datasets and Experimental Settings}
\label{sec:datasets}

\begin{table*}[ht]
        \centering
        \caption{Accuracy of recognition (\%) on NTU-60 and NTU-120 datasets. The evaluation is performed using cross-view and cross-subject settings on NTU-60 and cross-subject and cross-setup settings on NTU-120. *These values have not been reported in the state-of-the-art and the available codes have been used to obtain the recognition accuracy of these algorithms on NTU-120. }
        \label{tab:results_sota}
        \begin{tabular}{| c || c | c || c | c |}
        \hline
        \textbf{Method} & \multicolumn{2}{c ||}{\textbf{NTU-60 (\%)}} &  \multicolumn{2}{c |}{\textbf{NTU-120 (\%)}}\\
            \cline{2-5}
            & X-subject & X-view & X-subject & X-setup \\
            \hline
            SkeleMotion \cite{caetano2019skelemotion} & $76.5$ & $84.7$ & $67.7$ & $66.9$ \\
            \hline
            Body Pose Evolution Map \cite{liu2018recognizing} & $91.7$ & $95.3$ & $64.6$ & $66.9$ \\
            \hline
            Multi-Task CNN with RotClips \cite{ke2018learning} & $81.1$ & $87.4$ & $62.2$ & $61.8$ \\
            \hline
            Two-Stream Attention LSTM \cite{liu2017skeleton} & $76.1$ & $84.0$ & $61.2$ & $63.3$ \\
            \hline
            Skeleton Visualization (Single Stream) \cite{liu2017enhanced} & $80.0$ & $87.2$ & $60.3$ & $63.2$ \\

            \hline
            Multi-Task Learning Network \cite{ke2017new} & $79.6$ & $84.8$ & $58.4$ & $57.9$ \\
             \hline
            ST-GCN (10 blocks) \cite{yan2018spatial} & $81.5$ & $88.3$ & $72.4^*$ & $71.3^*$ \\
             \hline
             \hline
            \textbf{GVFE + ST-GCN w/ DH-TCN (4 blocks - ours)} & $ {79.1}$ & ${88.2}$  & ${73.0}$ & ${74.2}$\\
            \hline
            AS-GCN (10 blocks) \cite{li2019actional} & $\textbf{86.8}$ & $\textbf{94.2}$ & $77.7^*$ & $78.9^*$ \\
             \hline
             \hline
          

            \textbf{GVFE + AS-GCN w/ DH-TCN (4 blocks - ours)} & $ {85.3}$ & ${92.8}$ & $\mathbf{78.3}$ & $\mathbf{79.8}$\\
            \hline
          
        \end{tabular}
    \end{table*}

\textbf{NTU RGB+D Dataset (NTU-60)}: NTU RGB+D is a Kinect-acquired dataset which consists of $56,880$ videos. This dataset includes $60$ actions performed by $40$ subjects. The data are collected from $3$ different angles, at $\ang{-45}, \ang{0}$ and $\ang{45}$ with respect to the human body. In our experiments, we follow the same protocols (cross-view and cross-subject settings) proposed in~\cite{Shahroudy_2016_NTURGBD}.

\noindent \textbf{NTU RGB+D 120 Dataset (NTU-120)}: NTU RGB+D 120 Dataset extends the original NTU dataset by adding $60$ additional action classes to the existing ones and $66$ more subjects. The recording angles remain the same at $\ang{-45}, \ang{0}$ and $\ang{45}$ with respect to the human body, but more setups (height and distance) are considered ($32$ instead of $18$). We consider the same evaluation protocol ( cross-setup and cross-subject settings) suggested in~\cite{Liu_2019_NTURGBD120}.


\subsection{Implementation Details}
\label{sec:implementation}

The implementation of our approach is based on the PyTorch ST-GCN \cite{yan2018spatial} and AS-GCN \cite{li2019actional} codes. In both approaches, we include the GVFE module before the first ST-GCN block and we replace the temporal convolutions of each block with the DH-TCN module. For the spatial GCN, we use the same parameters suggested in \cite{yan2018spatial}. The number of output channels in GVFE is set to $C_{out}=8$ and we use $N=2$ hierarchical modules in DH-TCN. The temporal window of the DH-TCN module is set to $T_w=9$. The Stochastic Gradient Descent optimizer is used with a decaying learning rate of $0.01$. In contrast to~\cite{yan2018spatial, li2019actional} that makes use of $10$ ST-GCN or AS-GCN blocks, we use only $4$ blocks with $k \in \{1,...,4\}$.

\subsection{Results}
\label{sec:results}

\subsubsection{Comparison with state-of-the-art}

In this section, we compare our approach with recent skeleton-based methods, such as SkeleMotion~\cite{caetano2019skelemotion}, Body Pose Evolution Map~\cite{liu2018recognizing}, Multi-Task CNN with RotClips~\cite{ke2018learning}, Two-Stream Attention LSTM~\cite{liu2017skeleton}, Skeleton Visualization (Single Stream)~\cite{liu2017enhanced}, Multi-Task Learning Network~\cite{ke2017new} and more particularly with two the graph-based baselines namely ST-GCN~\cite{yan2018spatial} and AS-GCN~\cite{li2019actional}. GVFE and DH-TCN modules are incorporated in both ST-GCN~\cite{yan2018spatial} and AS-GCN~\cite{li2019actional} methods. The obtained accuracy of recognition on NTU-60 and NTU-120 datasets are reported in Table~\ref{tab:results_sota}.  

On NTU-120, we obtain the best accuracy of recognition of the state-of-the-art for both settings. Indeed, our approach used with AS-GCN (GVFE+AS-GCN w/ DH-TCN) reaches $78.3\%$ and $79.8\%$ for cross-subject and cross-setup settings, respectively. These positive results are also confirmed when testing our approach with ST-GCN (GVFE + ST-GCN w/ DH-TCN). Indeed, we improve the accuracy of the original ST-GCN by $0.6\%$ up to $2.9\%$.   

On NTU-60, the achieved scores are among the best of the state-of-the-art but remain slightly inferior to the original ST-GCN and AS-GCN (with respectively $79.1\%-88.2\%$ against $81.5\%-88.3\%$ and $85.3\%-92.8\%$ against $86.8\%-94.2\%$). Although being slightly inferior, it is important to highlight that only $4$ blocks are used in our case (against $10$ for ST-GCN and AS-GCN). The method based on Body Pose Evolution Map~\cite{liu2018recognizing} remains the best performing approach on NTU-60. 
However, this method registers an accuracy inferior to our approach by $13.7\%-12.9\%$, while the difference is less important on NTU-60 with only a gap of $6.4\%-2.5\%$ making our method more stable.

\subsubsection{Impact of the number of blocks}

As mentioned earlier, our approach utilizes only $4$ ST-GCN or AS-GCN blocks instead of $10$. For a fair comparison with the baselines, we also test ST-GCN~\cite{yan2018spatial} and AS-GCN~\cite{li2019actional} when using only $4$ blocks. The recognition accuracy of these experiments is reported in Table~\ref{tab:results_4blocks}.
Our method (GVFE + ST-GCN w/ DH-TCN) shows a significant performance boost in both settings of over $22\%$ compared to ST-GCN with $4$ blocks. Similarly, the recognition accuracy remains higher than the original AS-GCN compared to our method (GVFE + AS-GCN w/ DH-TCN). However, in this case, the accuracy boost is less impressive with an increase of $1.4\%$ for cross-subject settings and $0.4\%$ for cross-setup settings. This could be explained by the $7$ extra spatial-temporal convolutional blocks after the {\em maxPooling} layer in the AS-GCN network, which add more discriminative power to the full pipeline.


        
\begin{table}[ht!]
        \centering
        \caption{Accuracy of recognition (\%) using only $4$ ST-GCN or AS-GCN blocks on NTU-120 dataset for cross-subject and cross-setup settings. *These values are not reported in the state-of-the-art. Thus, the available codes have been used to obtain these results.}
        \label{tab:results_4blocks}
          \scalebox{0.76}{
        \begin{tabular}{| c | c | c | }
       
        \hline
            \textbf{Method} & \textbf{X-subject} & \textbf{X-setup} \\
             \hline
             ST-GCN (4 blocks)~\cite{yan2018spatial} & $45.3^*$ & $51.8^*$\\
             \hline
             \textbf{GVFE + ST-GCN w/ DH-TCN (4 blocks - ours)}& $\mathbf{73.0}$ & $\mathbf{74.2}$ \\
             \hline
             AS-GCN (4 blocks)~\cite{li2019actional} & $76.9^*$ & $79.4^*$ \\
             \hline
            \textbf{ GVFE + AS-GCN w/ DH-TCN (4 blocks - ours)} & $\mathbf{78.3}$ & $\mathbf{79.8}$ \\
             \hline
        \end{tabular}}
    \end{table}

\subsubsection{Ablation Study}
\label{sec:ablation_study}
To analyze the contribution of each component of our framework, an ablation study was conducted. For this purpose, we removed each time a component and report the obtained performance on both NTU-120 dataset for the cross-setup setting. The results are reported in Table~\ref{tab:results_ablation}. 

Our approach, which combines both the GVFE and the DH-TCN modules, achieves $74.2\%$ mean accuracy, which is higher by $22.4\%$ than the original ST-GCN approach with $4$ ST-GCN blocks. When using only the GVFE, the mean accuracy reaches $70.9\%$. We tested different configurations in this case, such as attaching a Rectified Linear Unit (ReLU) or a Batch Normalization Unit (BN). In both cases, the performance was degraded ($68.9\%$ and $66.7\%$, respectively), since these units distort the joint motion trajectories. 

Moreover, we conducted experiments by incorporating only the DH-TCN module. The mean accuracy, in this case, reached $68.3\%$, showing that GVFE and DH-TCN modules trained in an end-to-end manner can offer a significant performance boost.

\begin{table}[ht!]
        \centering
        \caption{Ablation study: accuracy of recognition (\%) on NTU-120 dataset for cross-setup settings using ST-GCN as a baseline. *These values are not reported in the state-of-the-art. Thus, the available codes have been used to obtain these results }
        \label{tab:results_ablation}
        \scalebox{0.85}{
        \begin{tabular}{| c | c | }
        
        \hline
            \textbf{Method} & \textbf{Accuracy (\%)} \\
             \hline
             ST-GCN (4 blocks)~\cite{yan2018spatial} & $51.8^*$\\
             \hline
             GVFE + ST-GCN (4 blocks) & $70.9$\\
             \hline
             ST-GCN w/ DH-TCN (4 blocks) & $68.3$\\
             \hline
             \textbf{GVFE + ST-GCN w/ DH-TCN (4 blocks - ours)} & $\mathbf{74.2}$\\
             \hline
        \end{tabular}}
    \end{table}

\subsubsection{Number of parameters and training time}

Although our method makes use of two additional modules compared to the baselines, the use of only $4$ blocks reduces the number of parameters. For instance, When using our method (GVFE + AS-GCN w/ DH-TCN) with $4$ blocks, the number of parameters drops from $7420696$ to $7370568$ compared to the original AS-GCN with $10$ blocks, while keeping almost the same accuracy on NTU-60 or even increasing it on NTU-120. Consequently, the training time is also reduced. As an example, on NTU-120 for cross-setup settings, our approach requires $24029$ seconds less than the original AS-GCN for training. 

\section{Conclusion}
\label{sec:conclusion}

In this paper, two novel modules for ST-GCN based methods have been proposed called GVFE and DH-TCN. These modules enable the reduction of the number of needed blocks and parameters while conserving almost the same or improving the recognition accuracy. Instead of relying on raw skeleton features such as skeleton joints, GVFE learns and generates graph vertex features in an end-to-end manner. To model simultaneously long-term and short-term dependencies, DH-TCN makes use of hierarchical dilated temporal convolutional layers. The relevance of these modules has been confirmed thanks to the performance achieved on two well-known datasets. Some future extensions are under consideration, such as applying a similar hierarchical model to replace the spatial graph convolutional layer. 

\section{Acknowledgements}
\label{sec:acknowledgements}
This work was funded by the European Union's Horizon 2020 research and innovation project STARR under grant agreement No.689947, and by the National Research Fund (FNR), Luxembourg, under the project C15/IS/10415355/3D-ACT/Bj\"{o}rn Ottersten. We would, also, like to thank Christian Hundt from NVIDIA AI Technology Center Luxembourg for his valuable input and fruitful discussions.

\bibliographystyle{IEEEbib}
\bibliography{icme2020template}

\end{document}